\documentclass{article} 
\usepackage{iclr2027_conference,times}
\iclrfinalcopy

\usepackage{amsmath,amsfonts,bm}

\def\eqref#1{equation~\ref{#1}}

\def\1{\bm{1}}

\DeclareMathAlphabet{\mathsfit}{\encodingdefault}{\sfdefault}{m}{sl}
\SetMathAlphabet{\mathsfit}{bold}{\encodingdefault}{\sfdefault}{bx}{n}

\usepackage{hyperref}
\usepackage{url}
\usepackage{booktabs}
\usepackage{graphicx}
\usepackage{multirow}
\usepackage{makecell}

\title{World Models Dream of Success:\\ Diagnosing and Repairing \\ Failure Insensitivity in Robot World Models}

\author{%
  Jiuyi Xu\textsuperscript{1}, \;
  Xiao Hu\textsuperscript{2}, \;
  Meida Chen\textsuperscript{3}, \;
  Peng Gao\textsuperscript{4}, \;
  Yang Ye\textsuperscript{2}, \;
  Yangming Shi\textsuperscript{1} \\[0.6ex]
  \textsuperscript{1}Colorado School of Mines \quad
  \textsuperscript{2}Northeastern University \\
  \textsuperscript{3}Institute for Creative Technologies, University of Southern California \\
   \textsuperscript{4}North Carolina State University
  }

\begin{document}

\maketitle
\lhead{Preprint.}

\begin{abstract}

Robot world models support policy evaluation, planning, and synthetic data generation, but these applications require predictions that distinguish successful actions from failures. 
Across four released checkpoints from two architecture families, we observe weak sensitivity to action changes and success-like predictions on verified failures. 
Although recent work incorporates failures into model training, which data can repair released checkpoints without changing their architecture or training objective still remains underexplored. 
To this end, we introduce \textbf{CureWM}, which constructs alternative actions from successful demonstrations across a severity grid, verifies their outcomes through execution in simulation or on hardware, and fine-tunes released models on the resulting failures and surviving successes alongside nominal demonstrations. 
This construction provides controlled action contrasts from shared starting contexts.
On 484 held-out LIBERO failure counterfactuals, optimism falls from 80\% after fine-tuning on the official data to 30--43\% across four independently fine-tuned CureWM models (38\% mean). 
In two separate evaluations on a physical robot arm, failure predictions scored as success-like by a latent-distance diagnostic decrease from 90\% after fine-tuning on successful demonstrations alone to 33\% with CureWM. 
With failure counts per task, successful replay data, and training budget matched, counterfactual failures yield a success--failure value gap of 0.124, compared with 0.014 for freshly collected on-policy failures.
These findings support execution-verified counterfactual replay for post-hoc repair and show why reduced optimism must be evaluated alongside success--failure discrimination.
Code is available at \url{https://github.com/jiuyixu25/CureWM}.

\end{abstract}

\section{Introduction} 
\label{sec:intro}   
Robot world models learn to predict how environments evolve in response to actions, from latent dynamics models to generative video simulators~\citep{dreamerv3,unisim,cosmos}.
In the robotic domain, their predictions support policy evaluation in imagination~\citep{ctrlworld}, action selection and planning~\citep{cosmospolicy}, and synthetic experience generation for policy learning~\citep{unisim,ctrlworld}. 
These applications require more than visually plausible futures: a model must distinguish actions that accomplish the task from those that fail. 
If failing actions receive high success values or generate success-like futures, decisions based on these predictions can favor actions that do not achieve the intended outcome. 

However, how to repair failure prediction in released robot world models still remains underexplored. 
MiraBench~\citep{mirabench} found optimism bias in action-conditioned predictions, but diagnosing such failure insensitivity does not tell us which training data could correct it. 
In particular, failed trajectories need not teach a model how alternative actions from the same context change the outcome.
This distinction matters for released checkpoints: simply adding failures may lower predicted success across both successful and failing actions, reducing the optimism without improving discrimination. 
Therefore, the question is \textit{which kind of data enable post-hoc repair of action-conditioned predictions, and how to distinguish that repair from a general shift toward pessimism.} 
Concurrent work, such as FACT~\citep{fact}, combines failed policy rollouts with action conditioning and failure-aware masking in a world-action model.
However, it does not directly address how to repair released checkpoints while retaining their architecture and training objective, or which failure data are most effective under these constraints. 

To fill these knowledge gaps, we introduce \textbf{CureWM}, motivated by the
observation that correctly predicting the outcome of an observed
action does not imply knowing how a different
action in the same context would change the outcome.
CureWM perturbs successful demonstrations across
a severity grid while preserving actions before
intervention, verifies outcomes through execution,
and retains failed and successful replays for
post-training.
The architecture and training objective remain unchanged.
We evaluate four released checkpoints from two architecture families, which are Cosmos Policy~\citep{cosmospolicy} and Ctrl-World~\citep{ctrlworld}, and test CureWM in simulation and on a physical robot arm. 
The ALOHA checkpoint is evaluated for diagnosis only (Appendix~\ref{app:aloha}).
Experiments on LIBERO~\citep{libero} and a physical
robot show improved success--failure discrimination.
Matched comparisons favor counterfactual failures
over fresh policy failures, while ablations examine
successful replays and verified outcome labels.
These results also show why reduced optimism alone
does not establish repair.
Our contributions are as follows:
\begin{enumerate}
  \item We introduce CureWM, which repairs released world models using execution-verified counterfactual replays without changing their architecture or training objective.

  \item We demonstrate repair in simulation and on
    a physical robot, including improved
    success--failure discrimination and failure detection.
    Matched LIBERO comparisons show stronger discrimination
    than training on fresh policy failures.

  \item We show that reduced optimism alone can overstate repair, and evaluate success--failure discrimination together with false alarms on successful actions.
\end{enumerate}

\section{Related Work}
\label{sec:related}

\textbf{World models and action-conditioned reliability.}
Cosmos Policy~\citep{cosmospolicy} predicts actions, future observations, and values for control and planning, while Ctrl-World~\citep{ctrlworld} supports policy evaluation and synthetic trajectory generation through action-conditioned video prediction.
MiraBench~\citep{mirabench} evaluates physics adherence, action following, and optimism bias, and provides the perturbation families used in our experiments. 
WorldArena~\citep{worldarena} likewise distinguishes perceptual quality from functional utility. 
We study post-hoc repair.
Our support argument relates to causal confusion
in imitation learning~\citep{causalconfusion},
but does not identify a causal mechanism inside
the model.

\textbf{Failure and counterfactual supervision.}
Learning from unsuccessful behavior and constructed examples has been widely studied: Dex-Net~\citep{dexnet2} learns grasp robustness from synthetic examples with analytic labels,
HER~\citep{her} reuses unsuccessful attempts through goal relabeling, 
and CoDA~\citep{coda} constructs counterfactual transitions using locally factored dynamics.
AHA~\citep{aha} synthesizes manipulation failures to train a vision-language model for failure detection and natural-language reasoning. 
Concurrent FACT~\citep{fact} incorporates failed rollouts into world-action model training.
We do not claim novelty in adding failure examples.
CureWM combines controlled perturbations, execution-verified outcomes, and successful replay partners to repair released checkpoints with their original architecture and objective.
Furthermore, matched comparisons test which data choices improve repair.

\textbf{Conservative values, failure detection, and predictive repair.}
Offline RL addresses unreliable extrapolation through data-constrained action selection \citep{bcq}, conservative value learning \citep{cql}, and uncertainty-aware penalties or pessimistic dynamics \citep{mopo,morel}.
These methods motivate caution about unsupported actions, but conservative decision-making and accurate outcome discrimination are distinct objectives.
Our label ablation illustrates that reduced optimism
  can accompany underestimated success predictions.
  It does not evaluate those offline RL algorithms. 
Foresight~\citep{foresight} instead learns failure detection from action-conditioned world-model representations. 
CureWM updates the world model's own predictions.
We assess whether it better distinguishes successful
from failing actions, rather than simply assigning
lower success values to both.

\section{Method}
\label{sec:method}

\subsection{Failure Insensitivity}
\label{sec:problem}

\begin{figure}[t]
  \centering
  \includegraphics[width=\linewidth]{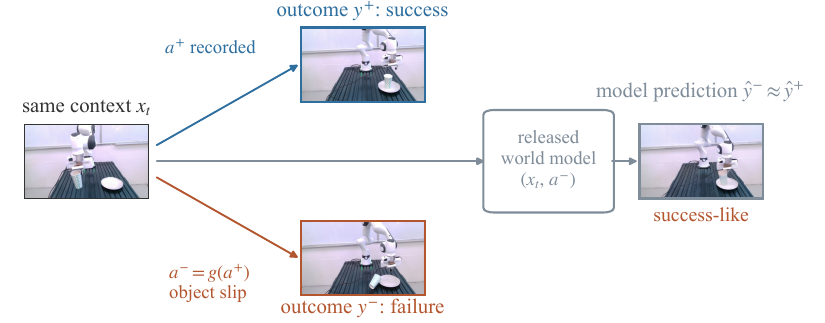}
  \caption{\textbf{Failure insensitivity on a held-out
  Franka counterfactual.}
  From the same starting context, the original action
  sequence succeeds and a perturbed sequence fails
  under physical execution, yet the model predicts
  a success-like future for the failing sequence.}
  \label{fig:insensitivity}
  \vspace{-5mm}
  \end{figure}

Let $x_t$ denote the observation history and task context,
and let $a^+_{t:T}$ and $a^-_{t:T}$ denote action sequences
that succeed and fail from the same starting context.
We call a model \emph{failure-insensitive} when its
predictions fail to distinguish these outcomes. Figure~\ref{fig:insensitivity} illustrates this behavior.
This differs from optimism alone: a model can assign
similar values to successful and failing actions even
when neither value is high.
We therefore evaluate both predictions on failing
actions and discrimination between successful and
failing actions.

\subsection{Measuring Failure Sensitivity}

\paragraph{Value prediction.}
For a model predicting native value predictions
$V_\theta(x,a)$, we measure the mean
\emph{success--failure value gap} over $N$ action pairs:
\begin{equation}
\Delta_{\mathrm{SF}}
=
\frac{1}{N}\sum_{i=1}^{N}
\left[
V_\theta(x_i,a_i^+)
-
V_\theta(x_i,a_i^-)
\right].
\label{eq:success-failure-gap}
\end{equation}
A positive gap means successful actions receive higher
values on average.
We report both values to check whether improved
discrimination comes at the cost of underestimating
successful actions. 
We use 0.5 as a fixed diagnostic threshold for the
native value prediction. We do not assume that this
value is a calibrated success probability.
False-positive rate measures successful replays
labeled as failures by this threshold.
We define \emph{value optimism} as the fraction of
verified failing sequences with
$V_\theta(x_i,a_i^-)>\tau$, using $\tau=0.5$ on LIBERO.
Including successful perturbed replays, we also report
AUROC~\citep{fawcett2006roc} and the fraction of successful
replays incorrectly predicted to fail.

\paragraph{Video prediction.}
Let $\hat z_T(a)$ be the predicted image representation,
and let $y_T^+$ and $y_T^-$ be recorded successful and
failed outcomes for the corresponding action pair.
With the frozen encoder $\phi$, we define
\begin{equation}
s(a)
=
d\!\left(\hat z_T(a),\phi(y_T^-)\right)
-
d\!\left(\hat z_T(a),\phi(y_T^+)\right),
\label{eq:video-margin}
\end{equation}
where $d$ is mean squared distance between image
representations, averaged across the camera views
used by the model.
Positive scores indicate predictions closer to the
successful reference.
We report \emph{visual optimism} which is the fraction of
failing sequences with $s(a^-)>0$, 
the mean failure-action score,
the mean paired difference
$\Delta_{\mathrm{SF}}=\frac{1}{N}\sum_i
[s(a_i^+)-s(a_i^-)]$, 
and pair-ranking accuracy, the fraction with
$s(a_i^+)>s(a_i^-)$.
These are offline diagnostics requiring recorded
reference outcomes.

\subsection{CureWM}
\label{sec:counterfactuals}

\begin{figure}[t]
  \centering
  \includegraphics[width=\linewidth]{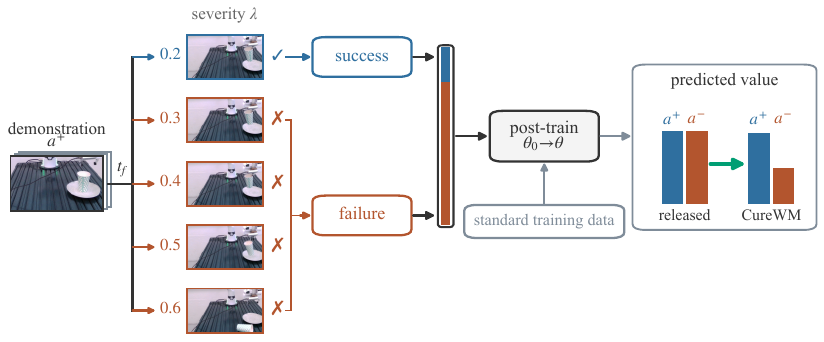}
  \caption{\textbf{The overview of CureWM.} We perturb successful demonstrations at different severities,
  verify outcomes through execution, and use both successful
  and failing replays for post-training alongside the original
  training data.
  Severity values, outcome assignments, and predicted-value
  bars are used for illustration rather than measured results.
  }
  \label{fig:curewm}
  \vspace{-5mm}
\end{figure}

CureWM constructs execution-verified action alternatives
for post-training (Figure~\ref{fig:curewm}).
It is motivated by the observation that, when only one
action is observed per context, fitting its outcome
does not require distinguishing the consequences of
alternative actions.
Proposition~1 in Appendix~\ref{app:theory} formalizes
this observation.

\paragraph{Controlled counterfactual construction.}
  We represent a successful demonstration as
  $\xi^+=(x_0,o_{0:T},a^+_{0:T-1},\ell=1)$,
  where $x_0$ is the initial context and $\ell$ indicates
  task success.
  We identify approach, grasp, carry, and place phases
  and select an intervention time $t_f$ for perturbation
  family $f$.
  Following MiraBench~\citep{mirabench}, we use approach
  overshoot, carry slip, contact oscillation, insufficient
  grip, premature release, and wrist tilt where applicable.
  For severity $\lambda\in\Lambda_f$, we construct
  \begin{equation}
  a^{f,\lambda}_{0:T-1}
  =
  g_{f,\lambda}\!\left(a^+_{0:T-1}\right),
  \qquad
  a^{f,\lambda}_t=a^+_t
  \quad \text{for }t<t_f.
  \label{eq:counterfactual-construction}
  \end{equation}
  The severity grid $\Lambda_f$ produces multiple action
  alternatives while preserving the actions before
  intervention.
  We do not label these alternatives as failures
  before execution.

  \paragraph{Execution-based outcome verification.}
  We execute each modified sequence from the corresponding
  initial context in simulation or on hardware, recording
  its observations, executed actions, and task outcome
  $\ell\in\{0,1\}$.
  A perturbation family enters training only when its
  empirical failure rate satisfies
  $\hat p_f(\lambda_{j+1})\geq\hat p_f(\lambda_j)$
  for adjacent increasing severities.
  This condition applies to aggregate failure rates,
  not to every demonstration individually.
  We retain successful and failing replays as
  $\mathcal D_{\mathrm{rep}}^+$ and
  $\mathcal D_{\mathrm{rep}}^-$, respectively.
  Individual families need not produce both outcomes
  at the tested severities.

  \paragraph{Post-training with verified trajectories.}
  We convert replays to the released model's training
  format using $\mathcal C$ and combine them with its
  base training data:
  \begin{equation}
  \mathcal D_{\mathrm{cure}}
  =
  \mathcal D_{\mathrm{base}}
  \uplus
  \mathcal C(\mathcal D_{\mathrm{rep}}^+)
  \uplus
  \mathcal C(\mathcal D_{\mathrm{rep}}^-).
  \label{eq:curewm-data}
  \end{equation}
  Here $\mathcal D_{\mathrm{base}}$ contains the
  demonstrations and official rollouts used for the
  corresponding model, and $\uplus$ combines the data
  without removing duplicates.
  For models with a value head, failed replays receive
  zero value targets.
  Successful replays retain the original labeling rule,
  which discounts the final success reward according
  to the remaining steps.
  For video prediction, supervision comes from the
  observations recorded during execution.

  Starting from released parameters $\theta_0$, we
  fine-tune the full prediction network using its
  original objective:
  \begin{equation}
  \min_\theta\;
  \mathbb E_{u\sim q_{\mathrm{cure}}}
  \left[\mathcal L_{\mathrm{native}}(\theta;u)\right],
  \qquad
  \theta\leftarrow\theta_0
  \quad\text{at initialization},
  \label{eq:curewm-objective}
  \end{equation}
  where $u$ is a training example and $q_{\mathrm{cure}}$
  follows the model's existing sampling procedure on
  $\mathcal D_{\mathrm{cure}}$.
  Successful and failing trajectories are related
  through their construction, but are not supplied as
  explicit training pairs.
  We introduce neither a new prediction head nor
  a pairwise loss.
  Training and sampling settings are provided in
  Appendix~\ref{app:repro}.

\section{Simulation Experiments}
\label{sec:sim}

\subsection{LIBERO}
\label{sec:libero}

\subsubsection{Evaluation Setup}

The baseline and CureWM use official data from
all four LIBERO suites; CureWM additionally uses
counterfactual replays from LIBERO-Goal.
Goal evaluation contains 484 verified failures
and 1,196 successful perturbed replays from two
independently generated sets using the training
perturbation families and severity grid.
Their source demonstrations are excluded from
counterfactual construction but mostly present
in the official data.
The second set was generated after the compared
models finished training.
Spatial, Object, and LIBERO-10 test transfer without
counterfactual training replays from those suites.
Robot task success uses 500 episodes per suite
and model.

\subsubsection{Results}

\begin{table}[t]
  \centering
  \caption{\textbf{LIBERO prediction and robot task success.}
  Baseline and CureWM use the same fine-tuning steps.
  Suite counts indicate failed/successful held-out replays.
  False-positive rate measures successful replays assigned
  a value at or below 0.5.
  Means weight the four suites equally.}
  \label{tab:libero-results}
  \label{tab:master}
  \small
  \setlength{\tabcolsep}{4pt}
  \renewcommand{\arraystretch}{1.12}
  \begin{tabular}{@{}llccccc@{}}
  \toprule
  Suite
  & Model
  & \makecell[c]{Value optimism\\$\downarrow$}
  & $\Delta_{\mathrm{SF}}\uparrow$
  & AUROC $\uparrow$
  & \makecell[c]{False-positive\\rate $\downarrow$}
  & \makecell[c]{Task success\\$\uparrow$} \\
  \midrule

  \multirow{3}{*}{\makecell[l]{Goal\\(484/1196)}}
  & Released & 79.13 & $+0.002$ & 0.497 & 30.5 & 98.4 \\
  & Baseline & 79.55 & $+0.003$ & 0.495 & 31.0 & 96.4 \\
  & CureWM   & 30.17 & $+0.282$ & 0.743 & 38.0 & 94.8 \\
  \midrule

  \multirow{3}{*}{\makecell[l]{Spatial\\(367/443)}}
  & Released & 65.67 & $-0.014$ & 0.518 & 36.6 & 98.4 \\
  & Baseline & 64.03 & $-0.011$ & 0.522 & 36.3 & 95.6 \\
  & CureWM   & 47.68 & $+0.081$ & 0.629 & 35.0 & 96.2 \\
  \midrule

  \multirow{3}{*}{\makecell[l]{Object\\(286/524)}}
  & Released & 54.55 & $+0.004$ & 0.464 & 57.4 & 99.8 \\
  & Baseline & 53.85 & $+0.004$ & 0.466 & 57.4 & 99.0 \\
  & CureWM   & 37.41 & $+0.086$ & 0.587 & 59.4 & 99.8 \\
  \midrule

  \multirow{3}{*}{\makecell[l]{LIBERO-10\\(359/331)}}
  & Released & 2.51 & $-0.001$ & 0.479 & 99.7 & 98.0 \\
  & Baseline & 2.51 & $-0.001$ & 0.483 & 99.1 & 96.2 \\
  & CureWM   & 2.51 & $+0.003$ & 0.493 & 99.4 & 95.4 \\
  \midrule

  \multirow{3}{*}{Mean}
  & Released & 50.47 & $-0.002$ & 0.490 & 56.1 & 98.65 \\
  & Baseline & 49.99 & $-0.001$ & 0.492 & 56.0 & 96.80 \\
  & CureWM   & 29.44 & $+0.113$ & 0.613 & 58.0 & 96.55 \\
  \bottomrule
  \end{tabular}
  \vspace{-5mm}
  \end{table}

We compare CureWM with the released model and
a baseline fine-tuned on official data for the
same number of steps. 
The detailed numerical results are presented in Table~\ref{tab:libero-results}.
On LIBERO-Goal, CureWM reduces value optimism
from the baseline's 79.55\% to 30.17\%,
increases $\Delta_{\mathrm{SF}}$ from 0.003 to 0.282,
and improves AUROC from 0.495 to 0.743.
However, the proportion of successful replays
incorrectly predicted to fail increases from
31.0\% to 38.0\%.
CureWM improves discrimination, but the fixed-threshold
detector predicts more successful replays as failures.
Without counterfactual training replays from
LIBERO-Spatial or LIBERO-Object, CureWM improves their AUROC
from 0.522 and 0.466 to 0.629 and 0.587,
respectively, alongside larger value gaps and
lower optimism.
On LIBERO-10, AUROC remains near chance at 0.493.
At the fixed threshold of 0.5, low optimism of 2.51\%
coexists with a 99.4\% false-positive rate.
Thus, low optimism does not indicate reliable failure
detection on this suite.
Mean task success is 98.65\% for Released,
96.80\% for the baseline, and 96.55\% for CureWM,
whose changes relative to the baseline vary by suite
(Table~\ref{tab:libero-results}).

Additional experiments assess training variability
and discrimination within perturbation types.
Across four independently fine-tuned CureWM models,
Goal value optimism ranges from 30.2\% to 43.0\%,
averaging 37.9\%
(Appendix~\ref{app:seeds}).
Within 21 eligible perturbation-family--severity
groups, mean AUROC increases from 0.537 to 0.714.
The improvement therefore cannot be explained
by perturbation type or severity alone
(Appendix~\ref{app:severity-discrimination}).
Experiments excluding individual perturbation types
from counterfactual training show uneven transfer
(Appendix~\ref{app:lofo}).

\subsection{RoboCasa}
\label{sec:robocasa}

\subsubsection{Evaluation Setup}

We evaluate failure detection on 111 successful and
114 failing drawer and door manipulation rollouts
excluded from training, reporting AUROC from predicted
task-success values at 75\% of episode length.
Separately, we measure $\Delta_{\mathrm{SF}}$ on
120 examples with modified actions from official
rollouts.
These actions are not re-executed, so this evaluation
measures prediction sensitivity rather than detection
of verified failures.
Robot task success is evaluated on OpenDrawer and
CloseDoubleDoor from RoboCasa~\citep{robocasa} over 30 episodes per task and model,
using three evaluation seeds with ten episodes each.

\subsubsection{Results}

\begin{table}[t]
\centering
\caption{\textbf{RoboCasa prediction sensitivity and task success.}
All $\Delta_{\mathrm{SF}}$ values use the same 120 modified-action
examples, whose modified actions are not re-executed.
AUROC is held-out articulated failure detection on 225 rollouts
(111 successes, 114 failures) scored at 75\% of episode length.
Task success is out of 30 episodes (three evaluation seeds,
ten episodes each). Expanded replays add drawer and door tasks.
Masking removes action-prediction loss on failed replays.}
\label{tab:robocasa-results}
\small
\setlength{\tabcolsep}{5pt}
\renewcommand{\arraystretch}{1.15}
\begin{tabular}{@{}lrrrrr@{}}
\toprule
Training data / setting
& Steps
& AUROC $\uparrow$
& $\Delta_{\mathrm{SF}}\uparrow$
& OpenDrawer
& CloseDoubleDoor \\
\midrule
Released
& --- & 0.634 & 0.0170 & 28/30 & 30/30 \\
\midrule
Grasp replays
& 5k & 0.674 & 0.0672 & 17/30 & 23/30 \\
Grasp replays
& 20k & 0.677 & 0.0342 & 26/30 & 29/30 \\
\midrule
Expanded replays
& 5k & 0.839 & 0.0302 & 25/30 & 26/30 \\
Expanded replays
& 20k & 0.780 & 0.0193 & 25/30 & 28/30 \\
\midrule
Grasp replays, action loss masked
& 5k & 0.662 & 0.0308 & 20/30 & 30/30 \\
\bottomrule
\end{tabular}
\vspace{-6mm}
\end{table}

Expanded training achieves the highest held-out
detection AUROC at 5,000 steps: 0.839 versus
0.634 for the released model
(Table~\ref{tab:robocasa-results}).
Grasp-focused training instead produces the largest
value gap, 0.0672, but reduces OpenDrawer and
CloseDoubleDoor success from 28/30 and 30/30
to 17/30 and 23/30.
Prediction sensitivity, failure detection, and
task success therefore do not improve together.

Adding drawer and door replays improves detection
and task success relative to grasp-focused training
at 5,000 steps, while changing both coverage and
data quantity.
Its task-success advantage does not persist at
20,000 steps.
Masking failed-action prediction loss also improves
task success at 5,000 steps, but reduces the
value gap and does not improve detection AUROC
relative to unmasked training.
Further comparisons are provided in
Appendix~\ref{app:robocasa}.

\subsection{Ctrl-World}
\label{sec:ctrlworld}

\subsubsection{Evaluation Setup}

Starting from the same LIBERO-adapted Ctrl-World
checkpoint, we compare CureWM with fine-tuning on
unperturbed successful demonstrations.
We evaluate predictions over four interaction
steps (16 raw frames) using
Equation~\ref{eq:video-margin} and frozen SVD~\citep{svd} VAE
representations from three camera views.
The evaluation contains 30 action pairs, including
ten excluded from counterfactual training.
The full set partially overlaps training.

\subsubsection{Results}

\begin{table}[t]
  \centering
  \caption{
  Ctrl-World: full-pool and held-out evaluation.
  }
  \label{tab:ctrlworld-heldout}
  \small
  \setlength{\tabcolsep}{7pt}
  \renewcommand{\arraystretch}{1.1}
  \begin{tabular}{@{}lrrrr@{}}
      \toprule
      & \multicolumn{2}{c}{Full pool: 30 pairs}
      & \multicolumn{2}{c}{Held-out subset: 10 pairs} \\
      \cmidrule(lr){2-3}
      \cmidrule(l){4-5}
      Model
      & Optimism & Mean $s(a^{-})$
      & Optimism & Mean $s(a^{-})$ \\
      \midrule
      Adapted
      & 27/30 & +0.080
      & 9/10 & +0.076 \\

      Control
      & 26/30 & +0.069
      & 8/10 & +0.071 \\

      CureWM
      & 21/30 & +0.016
      & 6/10 & +0.005 \\
      \bottomrule
  \end{tabular}
  \vspace{-5mm}
\end{table}

On the ten held-out pairs, CureWM reduces visual
optimism from 8/10 to 6/10 and the mean failure-action
score from 0.071 to 0.005
(Table~\ref{tab:ctrlworld-heldout}).
The paired analysis supports the score reduction,
but the optimism-rate difference is not statistically
established.
Full-set results provide supporting evidence
because of their training overlap.
A second independently fine-tuned model improves
the full-set score in the same direction, without
a statistically established difference
($p\approx .18$).
Protocols and paired analyses are provided in
Appendix~\ref{app:ctrlworld}.

\section{Repair on a Physical Robot}
\label{sec:real}

\subsection{Evaluation Setup}

We evaluate CureWM on a Franka Research 3 using
DROID-pretrained Ctrl-World~\citep{droid,ctrlworld}.
Cup pick-and-place is the primary repair task.
Cube picking tests task-specific repair and transfer
from cup-task repair.
Replay outcomes are verified through physical execution.
After 1,400 steps of scene adaptation using successful
demonstrations, we fine-tune CureWM on 90 failed
replays, 24 successful wrist-tilt replays, and
31 unperturbed successful trajectories.
The baseline uses only the same 31 unperturbed
trajectories.
Both receive 3,600 additional training steps,
matching training duration but not distinct-data counts.
Successful perturbed replays come only from wrist tilt and 
the other three tested perturbation types produce
no successful replays at the evaluated severities.
Data collection and splits are detailed in
Appendix~\ref{app:hardware-pools}.
Each of two hardware evaluations contains 15 failing
action sequences paired with their original successful
sequences from ten held-out demonstrations.
These demonstrations and their perturbed replays
are excluded from training.
We measure visual optimism and success--failure
discrimination using Equation~\ref{eq:video-margin}.
We also measure failure-detection AUROC using distance
to the image recorded at the end of each successful
execution.
Both diagnostics require recorded reference outcomes.

\subsection{Results}

\begin{table}[t]
\centering
\caption{Cup pick-and-place evaluation on Franka Research 3. AUROC uses both successful and failed reference outcomes.}
\label{tab:real}
\small
\setlength{\tabcolsep}{5pt}
\renewcommand{\arraystretch}{1.1}
\begin{tabular}{@{}lcccc@{}}
  \toprule
  Model
  & Optimism
  & \makecell[c]{Pair-ranking\\accuracy (\%)}
  & $\Delta_{\mathrm{SF}}$
  & AUROC \\
  \midrule
\multicolumn{5}{@{}l}{\textit{First hardware evaluation}} \\
Adapted
& 13/15 & 80.0 & +0.077 & 0.75 \\
Baseline
& 15/15 & 66.7 & +0.066 & 0.65 \\
CureWM, model 1
& 6/15 & 100.0 & +0.251 & 0.95 \\
CureWM, model 2
& 6/15 & 100.0 & +0.271 & 1.00 \\
\midrule
\multicolumn{5}{@{}l}{\textit{Second hardware evaluation}} \\
Adapted
& 15/15 & 93.3 & +0.05 & 0.75 \\
Baseline
& 12/15 & 66.7 & +0.06 & 0.70 \\
CureWM, model 1
& 4/15 & 100.0 & +0.19 & 0.97 \\
CureWM, model 2
& 6/15 & 93.3 & +0.17 & 0.96 \\
\bottomrule
\end{tabular}
\end{table}

\begin{figure}[t]
\centering
\includegraphics[width=\linewidth]{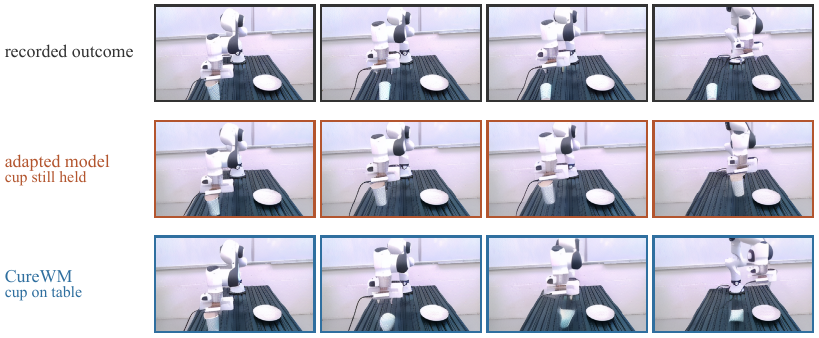}
\caption{\textbf{Predicted futures for a failing cup-task
action sequence.}
Columns show successive time steps.
Rows show recorded execution, the scene-adapted model,
and CureWM.
CureWM predicts the observed release;
the adapted model predicts continued grasping.}
\label{fig:hardware-predictions}
\vspace{-5mm}
\end{figure}

After success-only scene adaptation, 65 of 71 verified
cup-task failures receive success-like predictions
under the latent-distance diagnostic.
This is not a human-judged success rate, and most
of these examples subsequently enter repair training.
Repair is evaluated separately on held-out demonstrations.
In the first evaluation, CureWM reduces visual optimism
from the baseline's 15/15 to 6/15
(Table~\ref{tab:real}).
The mean failure-action score changes from
$+0.184$ to $-0.027$, while the success-action score
changes from $+0.249$ to $+0.224$.
The larger change therefore occurs on failing actions.
Before the second evaluation, all models receive
the same additional adaptation on successful
demonstrations and pass the rendering-quality check
(Appendix~\ref{app:hardware}).
Across both evaluations, visual optimism decreases
from 27/30 to 10/30, and failure-detection AUROC
using successful outcomes as references increases
from the baseline's 0.67 to 0.90.
Adding only the 24 successful wrist-tilt replays
to the baseline leaves first-evaluation optimism
at 15/15, although its mean failure-action score
decreases from 0.184 to 0.138.
These additional successful replays do not reproduce
CureWM's improvement.
Figure~\ref{fig:hardware-predictions} illustrates
the change in generated futures for a failing
cup-task action sequence.

Three non-author raters assessed blinded predictions
from the first evaluation.
For the 15 failing actions, majority judgments
classify 8/15 adapted-model predictions and 0/15
CureWM predictions as successful, compared with
13/15 and 6/15 under the distance-based diagnostic.
Human judgments support the direction of improvement
while showing that the diagnostic more frequently
classifies predictions as successful (Appendix~\ref{app:human}).
On cube picking, task-specific repair reduces the
mean failure-action score relative to the baseline
by 0.025 on 29 held-out pairs from ten scenes.
Applying cup-task repair before cube-scene adaptation
instead gives a difference of $-0.007$.
The results support task-specific repair but do not
establish transfer from cup to cube picking
(Appendix~\ref{app:cube}).
\section{Discussion}
\label{sec:analysis}

\subsection{Counterfactual Failures versus Policy Failures}
\begin{table}[t]
  \centering
  \small
  \setlength{\tabcolsep}{5pt}
  \renewcommand{\arraystretch}{1.12}
  \caption{\textbf{Failure sources and successful replays on LIBERO.}
  All models are evaluated on the same 484 held-out
  failing replays and their successful counterparts.
  The 7,500-step comparison matches failure counts
  and successful training replays.
  The 17,500-step comparisons report the primary models.
  The two groups use different failure pools and
  training budgets. $\Delta_{\mathrm{SF}}$ is computed before rounding.
  }
  \label{tab:onpolicyarms}
  \label{tab:partners}
  \begin{tabular}{@{}lrrrr@{}}
  \toprule
  Training setting
  & Value optimism (\%) $\downarrow$
  & Mean $V(a^-)$
  & Mean $V(a^+)$
  & $\Delta_{\mathrm{SF}}\uparrow$ \\
  \midrule
  Released model
  & 79.1 & 0.601 & 0.603 & $+0.002$ \\
  \midrule
  \multicolumn{5}{@{}l}{
  \textit{Matched failure-source comparison: 7,500 steps}} \\
  Official-data baseline
  & 79.3 & 0.594 & 0.597 & $+0.003$ \\
  Fresh policy failures
  & 66.9 & 0.545 & 0.559 & $+0.014$ \\
  Matched counterfactual failures
  & 46.9 & 0.463 & 0.587 & $+0.124$ \\
  \midrule
  \multicolumn{5}{@{}l}{
  \textit{Training-data comparisons: 17,500 steps}} \\
  Official-data baseline
  & 79.5 & 0.602 & 0.605 & $+0.003$ \\
  Own-failure up-sampling
  & 78.3 & 0.599 & 0.603 & $+0.004$ \\
  No-graded
  & 49.0 & 0.441 & 0.593 & $+0.152$ \\
  Other-demo
  & 60.1 & 0.495 & 0.603 & $+0.107$ \\
  CureWM
  & 30.2 & 0.300 & 0.582 & $+0.282$ \\
  \bottomrule
  \end{tabular}
  \vspace{-5mm}
  \end{table}

On LIBERO, we compare 140 fresh policy failures
  with 140 counterfactual failures, matching per-task
  failure counts, the same 580 successful replays,
  and 7,500 training steps.
  On 484 held-out counterfactuals, their value gaps
  are 0.014 and 0.124, respectively
  (Table~\ref{tab:onpolicyarms}).
  The comparison is separate from the 17,500-step
  experiments.
Of the selected policy failures, 69 have successful
  rollouts from the same task and initial-state index,
  but their actions differ from the first step.
  Counterfactual replay preserves actions before
  intervention.
  Failure type, timing, and coverage also differ,
  so shared prefixes are not established as the
  sole explanation.
Up-sampling official failures leaves optimism
  at 78.3\%, but does not match coverage, novelty,
  or repetition.
  Neither comparison establishes that policy failures
  are ineffective in general
  (Appendix~\ref{app:onpolicy}).

\subsection{Successful Replays from the Same Demonstrations}

The \emph{no-graded} variant removes the 580 successful
perturbed replays, while \emph{other-demo} replaces
them with successful replays from different demonstrations,
approximately matched by perturbation family and severity.
Both retain the 242 verified failures and original
demonstration data.
Across two independently fine-tuned models per variant,
value optimism is 43.4--49.0\% for no-graded and
44.8--60.1\% for other-demo, compared with
30.2--43.0\% across four full CureWM models.
These results support an additional contribution
from successful replays of the same demonstrations.
However, the smallest observed difference is only
0.4 percentage points, and differences in task
composition, sampling frequency, and training
variation limit attribution to demonstration
identity alone (Appendix~\ref{app:controls}).

\subsection{Lower Optimism Does Not Always Mean Better Predictions}

Fresh policy failures reduce optimism by 12.4
percentage points while lowering values for both
successful and failing actions
(Table~\ref{tab:onpolicyarms}).
Labeling all perturbed replays as failures, including
580 successful replays, also lowers training optimism
but reduces the success-side value to 0.147.
Its held-out value gap is 0.103 versus CureWM's 0.282
(Appendix~\ref{app:controls}).
These comparisons motivate evaluating discrimination
and fixed-threshold false-positive rates alongside
optimism.

\subsection{Limitations}

Only 53 of 484 LIBERO-Goal failures come from
  demonstrations absent from the shared official data.
  Transfer across perturbation types and objects
  is inconsistent, and the prediction horizon limits
  observable failure consequences.
  Cosmos Policy's video repair remains partial.
  Hardware evidence covers two objects on one robot,
  with small held-out samples and unequal
  training-data counts.

  CureWM requires executable counterfactual actions
  and verifiable outcomes; hardware detection also
  requires episode-specific references.
  Predictive gains do not establish improved control:
  RoboCasa exhibits a repair--control trade-off,
  best-of-$N$ selection shows no statistically
  established benefit, and policy ranking is untested.
\section{Conclusion}
\label{sec:conclusion}

We introduce CureWM, which repairs released robot world models using execution-verified counterfactual replays without changing their architecture or training objective.
Experiments in simulation and on hardware show improved success--failure discrimination.
Matched comparisons favor counterfactual failures over fresh policy failures at the tested budget, while ablations support retaining successful replays from the same demonstrations.
These results also show why lower optimism alone does not establish repair.
Broader transfer and improved robot control remain open challenges.

\subsection*{AI use statement}

We used generative AI (i.e., Claude and ChatGPT) tools to assist with drafting, revising, and polishing the manuscript, and to assist
in proving the mathematical claims presented in
Appendix~\ref{app:theory}. 
The authors reviewed the AI-assisted text for accuracy
and consistency with the experimental evidence.
We also checked the mathematical arguments, including
their assumptions, logical steps, and the scope of
their conclusions, and revised them where necessary.
We take responsibility for the final content of
this work, including all AI-assisted text,
mathematical claims, and proofs.

\subsection*{Reproducibility statement}

We provide anonymized code and analysis scripts
in the supplementary materials.
Section~\ref{sec:method} describes counterfactual
construction, execution-based outcome verification,
and post-training.
Appendix~\ref{app:repro} documents training settings,
sampling configuration, and evaluated checkpoints.
Data composition and evaluation splits are detailed
in Appendices~\ref{app:pools},
\ref{app:robocasa}, \ref{app:ctrlworld},
and~\ref{app:hardware-pools}.
Hardware evaluation, human assessment, and cube-task
protocols are provided in Appendices~\ref{app:hardware},
\ref{app:human}, and~\ref{app:cube}, respectively.
The assumptions and proofs underlying our theoretical
argument appear in Appendix~\ref{app:theory}.

\bibliography{iclr2027_conference}
\bibliographystyle{iclr2027_conference}

\appendix
\providecommand{\deltadisc}{\ensuremath{\Delta_{\mathrm{SF}}}}

\section{Extended Related Work}
\label{app:related}

\textbf{World models and their evaluation.}
Latent world models support learning through imagined interaction \citep{worldmodels,dreamerv3}, while video simulators extend prediction to visually rich environments \citep{unisim,cosmos,ctrlworld}.
Their potential use with generalist policies \citep{openvla,pi0,groot} makes action-conditioned reliability important.
MiraBench \citep{mirabench} provides the perturbation families used in our experiments. 
Complementary evaluations examine embodied executability \citep{robowmbench} and dynamic consistency \citep{dyncons}. 
Research on predictable world-model hallucinations connects diagnosis to data coverage and mitigation \citep{hallupreventable}. 
These works motivate evaluating consequences beyond visual plausibility. 
CureWM studies the response of released checkpoints to execution-verified counterfactual data.

\textbf{Failure data, detection, and recovery.}
Large manipulation datasets provide a broader data context \citep{agibotworld}. 
Failure judgments, detection, backup behavior, and counterfactual recovery are studied by AHA \citep{aha}, Foresight \citep{foresight}, AEGIS \citep{aegis}, and CoRe \citep{core}. Counterfactual failure synthesis also supports recovery learning \citep{dream2fix}.
Our intervention changes the world model's predictions through post-training. 
It is distinct from fitting a separate detector or selecting a recovery action, although these approaches can be complementary. 

\textbf{Relation to FACT and negative-data learning.}
FACT \citep{fact} combines failure data with world-action-model design and failure-aware masking. 
CureWM instead retains released architectures and their original objectives in its main treatment.
Our matched fresh-failure experiment compares two data constructions within this fixed recipe and it is not an evaluation of FACT.
More generally, learning from unsuccessful attempts, relabeled goals, and counterfactual augmentation has substantial work \citep{dexnet2,her,coda}. 
Our contribution concerns the construction and evaluation of post-hoc repair data, not the introduction of negative examples. 
The support argument is also related to concerns about causal confusion \citep{causalconfusion}, without identifying a particular mechanism in the trained network. 
Conservative offline learning provides a separate context for distinguishing lower values from better discrimination \citep{bcq,cql,mopo,morel}.

\section{Theoretical Motivation}
\label{app:theory}

\subsection{Setting and Scope}

Let $D=\{(x_i,a_i,y_i)\}_{i=1}^{n}$ contain contexts,
actions, and binary outcomes $y_i\in\{0,1\}$.
We consider a predictor with logit $f(x,a)$ and
logistic loss
\[
\ell(f,y)
=
-y\log\sigma(f)
-(1-y)\log(1-\sigma(f)),
\qquad
\sigma(f)=\frac{1}{1+e^{-f}}.
\]
Its empirical risk is
\[
\hat R_D(f)
=
\frac{1}{n}\sum_{i=1}^{n}
\ell(f(x_i,a_i),y_i).
\]
Let $\mathcal H_{\mathrm{state}}$ contain all
real-valued functions of $x$ alone.
This unrestricted class describes what the observed
data can distinguish. 
It is not an assumption about the ability or learned features of a particular model.
The logit $f$ is distinct from the video score
$s(a)$ in Equation~\ref{eq:video-margin}.
The following results concern a simplified
logistic objective, whereas the released models
use their original training objectives.
They do not establish optimization behavior
or held-out generalization.

\subsection{Single-Action Observations}

\paragraph{Proposition 1.}
Suppose every observed action satisfies
$a_i=\pi(x_i)$ for some function $\pi$.
For every predictor $f(x,a)$, there is a state-only
predictor $h_f(x)=f(x,\pi(x))$ such that
\[
\hat R_D(h_f)=\hat R_D(f).
\]
Thus, fitting the observed outcomes does not require
distinguishing alternative actions within the
same context.

\paragraph{Proof.}
For every observation,
$h_f(x_i)=f(x_i,\pi(x_i))=f(x_i,a_i)$.
The predictors therefore incur identical loss
on each observation and have equal empirical risk.
$\square$

The result allows an action-insensitive fit.
It does not show that a trained network adopts it.
Parameter sharing, regularization, and information
from related contexts may still support predictions
for unobserved actions.
Stochastic policy data need not satisfy the
single-action assumption, and the proposition
does not imply that policy failures are uninformative.

\subsection{Matched Actions with Different Outcomes}

\paragraph{Proposition 2.}
Suppose $D$ contains $m$ pairs
\[
\bigl\{(x_j,a_j^+,1),(x_j,a_j^-,0)\bigr\},
\qquad a_j^+\neq a_j^-,
\]
with each observation index used in at most one
counted pair.
Let $I_{\mathrm{rest}}$ contain the remaining
observation indices, and define
\[
\hat R_{\mathrm{rest}}^*
=
\inf_{h\in\mathcal H_{\mathrm{state}}}
\frac{1}{n}
\sum_{i\in I_{\mathrm{rest}}}
\ell(h(x_i),y_i).
\]
Then
\[
\inf_{h\in\mathcal H_{\mathrm{state}}}
\hat R_D(h)
\geq
\frac{2m}{n}\log 2
+
\hat R_{\mathrm{rest}}^*.
\]
For an action-conditioned predictor, the summed
loss on a pair with logit difference
$\Delta=f(x,a^+)-f(x,a^-)$ satisfies
\[
\ell(f(x,a^+),1)+\ell(f(x,a^-),0)
\geq
2\log\!\left(1+e^{-\Delta/2}\right).
\]

\paragraph{Proof.}
A state-only predictor assigns the same logit $u$
to both observations in a pair.
Their summed loss is
\[
\log(1+e^{-u})+\log(1+e^u),
\]
which is minimized at $u=0$ with value $2\log 2$.
Since counted pairs do not share observation indices,
these bounds can be summed without counting a
loss term twice.
Adding the lower bound for the remaining
observations proves the first inequality.

For a fixed difference $\Delta$, write the two
logits as $c+\Delta/2$ and $c-\Delta/2$.
Their summed loss is
\[
L(c)
=
\log(1+e^{-c-\Delta/2})
+
\log(1+e^{c-\Delta/2}).
\]
This function is convex and symmetric in $c$,
so its minimum occurs at $c=0$, giving
$2\log(1+e^{-\Delta/2})$.
$\square$

If labels are consistent at each context--action
input and the predictor can represent arbitrarily
large separating logits, the loss on matched
pairs can approach zero.
These conditions are not guaranteed for a
released model.
When counting the $m$ pairs, each observation
may be used at most once, even if a failed
observation could be paired with several successful
observations from the same context.

\subsection{Implications for CureWM}

The propositions motivate constructing action
alternatives with execution-verified different
outcomes under matched contexts.
They do not establish that successful perturbed
replays are necessary for repair:
original demonstrations may already provide
successful alternatives.
Additional successful replays are evaluated
empirically rather than justified by a necessity
claim.
Exact shared contexts are an idealization.
A common source demonstration or action prefix
does not by itself guarantee identical observed
states during replay, especially on hardware.
The argument also does not identify shared prefixes
as the sole cause of the observed improvements,
exclude useful learning from policy failures,
or guarantee transfer to new tasks.
Training-step comparisons at fixed data composition
do not test the bound's dependence on $m/n$.

\section{Additional Diagnosis on ALOHA}
\label{app:aloha}

We evaluate the released Cosmos Policy checkpoint
for ALOHA~\citep{aloha} on action sequences from
40 recorded episodes.
For each episode, we apply three action edits:
freezing the actions, reversing the action sequence,
and opening both grippers.
The reported changes in predicted task-success
values are below 0.01 in magnitude, indicating
limited sensitivity to these edits.
The modified actions are not re-executed, so their
outcomes are not verified.
This experiment measures sensitivity to action
changes, rather than prediction accuracy on confirmed
failures.
ALOHA is included for diagnosis only and we do not
evaluate CureWM repair on this checkpoint.

\section{LIBERO}
\label{app:libero}

\subsection{Training Data and Evaluation Splits}
\label{app:pools}

The official training data contain 2,000 records,
including 300 failed policy rollouts.
CureWM adds 822 counterfactual replay records:
242 verified failures and 580 successful replays.
These include twelve repeated successful records.
The failures comprise 23 approach-overshoot,
seven carry-slip, 32 contact-oscillation,
56 insufficient-grip, 67 premature-release,
and 57 wrist-tilt replays.
Using one-based task and demonstration indices,
counterfactual training uses demonstrations 1--3
per LIBERO-Goal task.
The first evaluation set uses demonstrations 4--6,
the second uses 7--9, and the other-demo comparison
uses demonstrations 10--12.
Sources whose original action sequence fails during
replay are skipped, leaving 27 training sources
and 56 evaluation sources.
The second evaluation set is generated after
the compared models finish training.
The evaluation contains 484 failures from 45
source demonstrations and 1,196 successful
perturbed replays.
All compared models use the same evaluation records.
The training-pool evaluations instead use two
groups containing 115 and 127 failures.
These are in-sample for CureWM and do not
establish held-out repair.
Most evaluation demonstrations also occur in
the shared official training data.
Three are absent: task 4/demonstration 6,
task 7/demonstration 5, and task 4/demonstration 9.
They contribute 20, 12, and 21 failures, respectively.
On these 53 failures, value optimism is
26.4\%, 28.3\%, and 11.3\% for the released model,
baseline, and CureWM.
With only three source demonstrations, this subset
does not isolate demonstration familiarity from
task and failure-type differences.
The LIBERO-Spatial, LIBERO-Object, and LIBERO-10 evaluations
contain 367/443, 286/524, and 359/331
failed/successful perturbed replays, respectively.
Their official data remain included in fine-tuning,
but their counterfactual replays do not enter training.
The suite means in Table~\ref{tab:libero-results}
weight the four suites equally.

\subsection{Training Configuration and Checkpoints}
\label{app:repro}
\label{app:seeds}

Cosmos Policy fine-tunes the full prediction network,
rather than a selected head or adapter.
The configuration sets
\texttt{demonstration\_sampling\_prob=0.5} and
\texttt{success\_rollout\_sampling\_prob=0.5}.
Successful and failing trajectories are sampled
through the existing training procedure, not supplied
as explicit positive--negative pairs.
Training-record counts therefore do not directly
specify sampling probabilities.
LIBERO uses batch size 4 and a default learning
rate of $10^{-5}$.
Primary comparisons use 17,500 steps because this
is the checkpoint retained for the primary CureWM
model within a 20,000-step schedule.
This evaluation step was not a prospectively
specified stopping rule.
Table~\ref{tab:runmap} reports the retained
checkpoints and corresponding results.

\begin{table}[htbp]
\centering
\small
\caption{Four independently fine-tuned CureWM models
on LIBERO.
Values are optimism percentages.
Training groups A and B contain 115 and 127 failures, while
held-out evaluation uses the same 484 failures
for every model.}
\label{tab:runmap}
\begin{tabular}{@{}lrrrr@{}}
\toprule
Model & Steps & Training A & Training B & Held-out \\
\midrule
Primary & 17,500 & 16.5 & 17.3 & 30.2 \\
Supplementary 1 & 20,000 & 28.7 & 29.9 & 43.0 \\
Supplementary 2 & 20,000 & 18.3 & 20.5 & 37.6 \\
Supplementary 3 & 17,500 & 21.7 & 32.3 & 40.9 \\
\bottomrule
\end{tabular}
\end{table}

LIBERO retains the native discounted-return supervision
with $\gamma=0.99$.
For each action pair, the probe selects the frame
immediately before the first action difference,
adjusting the index when necessary to retain a full
16-action chunk.
Both predictions use the same observation, task
instruction, query time, and generation seed.
The probe generates a future state under each chunk
and then queries its value.
The resulting value is not treated as a calibrated
probability of eventual task success.

The first model is designated primary and also
achieves the lowest observed held-out optimism.
The 17,500-step checkpoints for supplementary
models 1 and 2 were not retained.
Variation therefore combines training randomness
and training duration.
All four models are compared with the same primary
baseline, whose optimism is 79.5\%;
a second official-data baseline achieves 78.5\%.
The comparisons do not use four independently
seed-matched baselines.
At learning rate $5\times10^{-6}$, three additional
models reduce the training-pool optimism ranges
from 12.2 to 4.4 percentage points on group A
and from 15.0 to 5.5 on group B.
Mean optimism is approximately 2.5 points higher.
These are training-pool results, not additional
held-out repair estimates.
Closed-loop evaluation uses the released evaluation
script, action chunks of length 16, five action
denoising steps, and evaluation seed 195.
Each suite contains ten tasks with 50 episodes
per task, giving 500 episodes per model and suite.

For the other model settings, RoboCasa uses batch
size 2 and three camera views.
The Ctrl-World protocols use 1,400 adaptation steps
followed by 3,600 comparison steps.
Additional hardware adaptation and cube-task
evaluations are specified in
Appendices~\ref{app:hardware} and~\ref{app:cube}.
The submitted configurations and analysis scripts
provide the executable settings.

\subsection{Discrimination within Perturbation Types}
\label{app:severity-discrimination}

We evaluate the same 484 failures and 1,196
successful perturbed replays.
Table~\ref{tab:falsealarm} separates results by
perturbation type.
Improvement is concentrated in insufficient grip,
premature release, and wrist tilt.
Wrist tilt also produces the largest increase
in successful replays incorrectly predicted to fail.

\begin{table}[htbp]
\centering
\small
\caption{LIBERO-Goal discrimination by perturbation type.
AUROC and false-positive rates compare the fine-tuning
baseline with CureWM on identical replays.
False-positive rates are percentages of successful
replays assigned a value at or below 0.5.}
\label{tab:falsealarm}
\setlength{\tabcolsep}{4pt}
\begin{tabular}{@{}lrrcccc@{}}
\toprule
& & & \multicolumn{2}{c}{AUROC}
& \multicolumn{2}{c}{False-positive rate} \\
\cmidrule(lr){4-5}\cmidrule(l){6-7}
Perturbation & Failed & Successful
& Baseline & CureWM & Baseline & CureWM \\
\midrule
Approach overshoot & 33 & 247 & 0.88 & 0.87 & 14.6 & 19.0 \\
Carry slip & 12 & 268 & 0.68 & 0.70 & 27.6 & 29.5 \\
Contact oscillation & 74 & 206 & 0.52 & 0.51 & 34.5 & 37.9 \\
Insufficient grip & 122 & 158 & 0.45 & 0.86 & 43.7 & 44.3 \\
Premature release & 140 & 140 & 0.57 & 0.78 & 42.1 & 42.9 \\
Wrist tilt & 103 & 177 & 0.41 & 0.72 & 35.0 & 68.4 \\
\bottomrule
\end{tabular}
\end{table}

To control for severity as well as perturbation type,
we compute AUROC separately within each combination.
The summary equally weights the 21 groups containing
at least three successful and three failing replays.
It is 0.537 for both the released model and baseline,
and 0.714 for CureWM.
This differs from pooled AUROC and from an average
over perturbation types alone. 
Table~\ref{tab:sevauroc} reports AUROC at fixed
perturbation type and severity

\begin{table}[htbp]
\centering
\small
\caption{AUROC at fixed perturbation type and severity.
Each group contains 56 replays.
Dashes indicate no failures.
Severity 0.2 is omitted because every type contains
at most two failures at that severity.
Carry slip at severity 0.6 contains only two failures
and is excluded from the 21-group mean.
Means use values before rounding.}
\label{tab:sevauroc}
\setlength{\tabcolsep}{4pt}
\begin{tabular}{@{}llcccc@{}}
\toprule
Perturbation & Model & 0.4 & 0.6 & 0.8 & 1.0 \\
\midrule
\multirow{2}{*}{Approach overshoot}
& Released & 0.85 & 0.94 & 0.89 & 0.96 \\
& CureWM   & 0.85 & 0.95 & 0.88 & 0.95 \\
\addlinespace
\multirow{2}{*}{Carry slip}
& Released & --- & 0.55 & 0.72 & 0.57 \\
& CureWM   & --- & 0.56 & 0.75 & 0.57 \\
\addlinespace
\multirow{2}{*}{Contact oscillation}
& Released & 0.67 & 0.52 & 0.55 & 0.38 \\
& CureWM   & 0.66 & 0.52 & 0.56 & 0.39 \\
\addlinespace
\multirow{2}{*}{Insufficient grip}
& Released & --- & 0.36 & 0.29 & 0.36 \\
& CureWM   & --- & 0.58 & 0.88 & 0.93 \\
\addlinespace
\multirow{2}{*}{Premature release}
& Released & 0.55 & 0.38 & 0.37 & 0.36 \\
& CureWM   & 0.56 & 0.49 & 0.84 & 0.90 \\
\addlinespace
\multirow{2}{*}{Wrist tilt}
& Released & 0.44 & 0.42 & 0.37 & 0.31 \\
& CureWM   & 0.61 & 0.66 & 0.69 & 0.78 \\
\midrule
\multirow{3}{*}{Mean over 21 groups}
& Released & \multicolumn{4}{c}{0.537} \\
& Baseline & \multicolumn{4}{c}{0.537} \\
& CureWM   & \multicolumn{4}{c}{0.714} \\
\bottomrule
\end{tabular}
\end{table}

Failure counts at severities 0.4, 0.6, 0.8, and 1.0 are
$(7,6,10,8)$ for approach overshoot,
$(0,2,4,5)$ for carry slip,
$(8,13,21,30)$ for contact oscillation,
$(0,40,41,40)$ for insufficient grip,
$(21,39,39,40)$ for premature release, and
$(12,23,31,35)$ for wrist tilt.
The remaining replays in each group succeed.
Outcome labels and evaluation records are shared
across models.

\subsection{Leave-One-Perturbation-Type-Out Evaluation}
\label{app:lofo}

For each evaluated perturbation type, we remove
all its counterfactual training replays while
retaining official data and the other types.
Two independently fine-tuned models are evaluated
at 17,500 steps on failures of the excluded type
from training-pool groups A and B.
These excluded replays do not enter that model's
counterfactual training, but other replays from
the same source demonstrations can remain.
The experiment tests transfer across perturbation
types, not unseen source demonstrations. 
Table~\ref{tab:lofo} reports the results for each
excluded perturbation type.

\begin{table}[htbp]
\centering
\small
\caption{Value optimism after excluding a perturbation
type from counterfactual training.
Each result lists the released model followed by
two independently fine-tuned models.
}
\label{tab:lofo}
\setlength{\tabcolsep}{5pt}
\begin{tabular}{@{}lrrll@{}}
\toprule
Excluded type & $n_A$ & $n_B$
& A: Released / models 1, 2
& B: Released / models 1, 2 \\
\midrule
Carry slip & 1 & 6
& 100 / 0, 0 & 16.7 / 16.7, 16.7 \\
Contact oscillation & 13 & 19
& 92.3 / 53.8, 69.2 & 57.9 / 57.9, 57.9 \\
Insufficient grip & 31 & 25
& 80.6 / 45.2, 19.4 & 60.0 / 24.0, 40.0 \\
Premature release & 36 & 31
& 88.9 / 41.7, 55.6 & 51.6 / 22.6, 22.6 \\
Wrist tilt & 24 & 33
& 83.3 / 75.0, 75.0 & 57.6 / 69.7, 69.7 \\
\bottomrule
\end{tabular}
\end{table}

Insufficient grip and premature release show
partial transfer.
Contact oscillation is unchanged on B, while
A contains only 13 failures.
Wrist tilt improves slightly on A but worsens on B.
Carry slip has only one A failure and six B failures.
Approach overshoot is not evaluated in this experiment.

\subsection{Failure-Source and Label Comparisons}
\label{app:onpolicy}
\label{app:controls}

\paragraph{Fresh policy failures.}
We execute the released policy with stochastic
sampling for 16 passes over ten LIBERO-Goal tasks
and 50 initial states per task, totaling 8,000 episodes.
The collection yields 238 verified failures.
61.8\% occur on two tasks, while one task has
no observed failure.
Zero observed failures does not imply zero
failure probability.

For each task, we take the smaller of the available
policy-failure and counterfactual-failure counts.
This gives 140 distinct failures per source.
Each set is repeated to form 242 training records
and combined with the same 580 successful
counterfactual replays and 2,000 official records.
Both models train for 7,500 steps.
Thus distinct-failure counts, per-task counts,
training-record counts, successful replay data,
and training duration are matched.

Of the 140 selected policy failures, 69 have
a recorded success from the same task and
initial-state index.
None shares an exactly identical first executed
action with its successful counterpart.
This criterion requires numerical equality of
every action component but it does not test approximate
action similarity.
All collected policy failures reach the 300-step
limit, which does not identify when a failure-causing
event occurred.
These construction differences do not establish
a unique explanation for the matched comparison
in Table~\ref{tab:onpolicyarms}.

\paragraph{Existing policy failures.}
The own-failure comparison repeatedly samples
300 failures already present in the official
mixture to produce 822 additional training records.
Unlike the fresh comparison, it does not match
task coverage, novelty, or repetition.
It therefore tests this specific reuse of existing
failures rather than policy-failure learning
in general.

\paragraph{Successful replay partners.}
No-graded removes the 580 successful replay records.
Other-demo replaces them with 580 verified successes
from demonstrations 10--12 per task, approximately
matched by perturbation type and severity.
Both retain the 242 verified failures and official
training data.
The full other-demo training list contains
2,822 records.
These comparisons change training-data composition;
record counts alone do not guarantee equal exposure
under the sampling procedure.

\paragraph{Outcome verification.}
The unverified-label comparison assigns failure
labels to all perturbed replays, including the
580 successful records, while retaining the remaining
training procedure.
It achieves 57.9\% held-out optimism and a value
gap of 0.103, compared with CureWM's 30.2\%
and 0.282.

On training failures, its optimism is 10.3\%,
below CureWM's 16.9\%.
However, its mean value for successful actions
is only 0.147 and its value gap is 0.062.
Lower training optimism therefore overstates repair:
the model also lowers successful-action predictions.

\paragraph{Failed-action supervision.}
Masking only action-prediction loss on failed
LIBERO replays gives 32.9\% held-out optimism
and a value gap of 0.258, compared with
30.2\% and 0.282 without masking.
Substantial value-side repair remains without
supervising those failed actions.
This prediction comparison does not establish
a closed-loop benefit from masking.

\subsection{Value and Video Predictions}

Cosmos Policy's value and video predictions do not
improve equally.
On the first held-out set of 231 failures,
value optimism is 71.4\%, 72.7\%, and 28.1\%
for the released model, baseline, and CureWM,
while visual optimism is 58.0\%, 55.8\%, and 45.0\%.

Wrist tilt illustrates the remaining visual error:
among its 49 failures, baseline-to-CureWM value
optimism falls from 93.9\% to 24.5\%, but visual
optimism falls only from 100.0\% to 81.6\%.
The results support partial video repair, rather
than uniform improvement across prediction outputs.

\section{RoboCasa}
\label{app:robocasa}

\subsection{Training Data and Evaluation Sets}

The initial grasp-focused pool
  (the Grasp replays setting in
  Table~\ref{tab:robocasa-results})
  contains 1,291 official rollouts and 513 counterfactual
  replays, including 262 verified failures.
The expanded pool adds 456 drawer and door replays,
including 113 failures.
This changes both task coverage and data quantity.

We distinguish three prediction evaluations:
\begin{enumerate}
\item The 120 modified-action examples measure
$\Delta_{\mathrm{SF}}$ on states from official
rollouts.
Modified actions are not re-executed.
All value gaps in Table~\ref{tab:robocasa-results}
use this same evaluation.
\item The official-rollout evaluation contains
148 successes and 92 failures included in the
training mixture.
It provides supporting results, not held-out evidence.
\item The separate drawer and door evaluation
contains 111 successes and 114 failures excluded
from training.
All held-out AUROCs in
Table~\ref{tab:robocasa-results} use this set
at 75\% of episode length.
\end{enumerate}

Robot control is evaluated over three evaluation
seeds with ten episodes each, giving 30 episodes
per task and model.
Neither OpenDrawer nor CloseDoubleDoor is covered
by the initial counterfactual training pool.

\subsection{Training Duration and Action Supervision}

Table~\ref{tab:robocasa-results} reports the primary
comparisons.
At 20,000 steps, a second independently fine-tuned
model using the initial pool gives
$\Delta_{\mathrm{SF}}=0.0357$, compared with
0.0342 for the primary model.
Two models trained by up-sampling existing policy
failures give 0.0131 and 0.0274.

The action-loss comparison masks only action-frame
loss on failed rollouts, retaining their value
and future-state supervision.
The pool, initialization, training seed, batch size,
and learning-rate schedule remain fixed.
At 5,000 steps, this improves observed task success
but reduces the value gap from 0.0672 to 0.0308.
It supports failed-action supervision as one
contributor to control-performance loss without
identifying all contributing mechanisms.

The recorded paired comparison on 120 modified-action
examples gives $t=10.39$, $p<.001$, and $d_z=0.948$
for unmasked treatment relative to the released model,
with 109 positive differences.
Masked treatment gives $t=8.5$ and $d_z=0.78$,
with 100 positive differences.
The masked-minus-unmasked comparison gives $t=-17.4$.
These example-level tests do not account for
dependence among examples from the same trajectory
and do not test differences in robot task success.

\subsection{Additional Prediction and Control Results}

On the official-rollout evaluation, predictions
are assessed at 25\%, 50\%, and 75\% of episode
length.
At 75\%, recall at a 10\% false-positive rate
increases from 53\% to 70\%.
These rollouts occur in the training mixture,
so this result does not establish generalization
to unseen rollouts or a deployment operating point.

On coffee-mug manipulation, absent from the initial
counterfactual pool, the value gap changes from
$-0.006$ to $+0.058$ at 5,000 steps.
This measures sensitivity to modified actions,
not verified failure detection.
Value-guided best-of-$N$ selection provides no
statistically established improvement in robot
task success at the evaluated scale.

\section{Ctrl-World}
\label{app:ctrlworld}

\subsection{Video Evaluation and Data Splits}

The simulator evaluation uses the frozen SVD VAE
representation and three camera views.
Distances average squared differences over the
representation dimensions of all views.
Predictions are evaluated after four chained
interaction steps, corresponding to 16 raw frames
after the initial context.

The expanded evaluation contains 30 action pairs
from 14 source demonstrations.
Ten pairs from nine sources are excluded from
counterfactual training.
The full set partially overlaps training and
is reported separately from this held-out subset
in Table~\ref{tab:ctrlworld-heldout}.

Positive $s(a^-)$ indicates a prediction closer
to the recorded success than the recorded failure.
A reduction in mean $s(a^-)$ is not by itself
proof of more accurate generated video.
The diagnostic requires recorded outcome references.

\subsection{Paired Comparisons and Independent Training}

Paired comparisons evaluate the baseline and CureWM
on identical examples.
The fixed analysis in \texttt{cw\_paired\_ci.py}
resamples source demonstrations with replacement,
retains all pairs from each sampled source,
and recomputes the mean using the resulting
number of pairs.
It uses seed 0 and 20,000 resamples.

The mean baseline-minus-CureWM failure-score
difference is 0.053 on all 30 pairs and 0.066
on the ten held-out pairs.
On the held-out subset, three optimistic predictions
are corrected and one changes in the opposite
direction.
The paired analysis supports a score decrease,
while the thresholded optimism difference is not
statistically established.

A second independently fine-tuned CureWM model
gives a smaller full-set score improvement of
0.0144, with reported $p\approx.18$.
Its mean held-out failure-action score is 0.0641.
The independently trained models are not pooled
as repeated observations of the same model.

\subsection{Original 23-Pair Evaluation}

The original evaluation contains 23 pairs and
partially overlaps counterfactual training.
Visual optimism is 95.7\%, 91.3\%, and 60.9\%
for the adapted model, baseline, and CureWM.
Relative to the baseline, seven predictions are
corrected and none reversed, giving a two-sided
exact McNemar $p=0.015625$.
The optimism difference is $-30.4$ percentage points.

The second CureWM model's reported comparison on
this original set has $p\approx.026$.
This is a different evaluation from the expanded
30-pair set, on which its improvement is not
statistically established.
Results from the two sets are not interchangeable.

\section{Real Robot}
\label{app:real}

\begin{figure}[t]
\centering
\includegraphics[width=\linewidth]{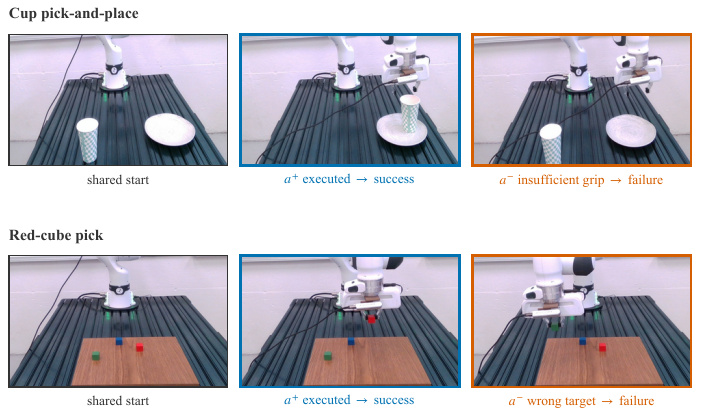}
\caption{\textbf{Hardware tasks and executed outcomes.}
Each row shows the starting scene, a successful
execution, and a failing execution.
Top: cup pick-and-place, with failure caused by
insufficient grip.
Bottom: red-cube picking, with failure caused by
grasping a distractor instead of the target cube.
All images show physical executions, not model predictions.}
\label{fig:realrobot-tasks}
\end{figure}

Figure~\ref{fig:realrobot-tasks} illustrates the
two hardware tasks and their execution-verified
successful and failing outcomes.

\subsection{Data Collection and Splits}
\label{app:hardware-pools}

Forty scripted cup pick-and-place demonstrations
produce 140 perturbed replays at severities
$\{0.6,0.8,1.0\}$:
107 verified failures and 33 successful replays.
The perturbation types are insufficient grip,
premature release, carry slip, and wrist tilt.
Only wrist tilt produces successful perturbed
replays at the tested severities.

Ten demonstrations and their 20 replays are
reserved for evaluation before training.
These replays contain 15 failures and five successes.
Neither the demonstrations nor their replays
enter training.
The complete split is shown in
Table~\ref{tab:hardware-splits}.

\begin{table}[htbp]
\centering
\small
\caption{Cup-task data split.
The 31 unperturbed training trajectories contain
28 demonstrations and three additional replays.
Successful perturbed replays all use wrist tilt.}
\label{tab:hardware-splits}
\begin{tabular}{@{}lrrr@{}}
\toprule
Data & Training & Validation & Evaluation \\
\midrule
Failed perturbed replays & 90 & 2 & 15 \\
Successful perturbed replays & 24 & 4 & 5 \\
Unperturbed successful trajectories & 31 & 2 & 10 \\
\bottomrule
\end{tabular}
\end{table}

After 1,400 steps of success-only scene adaptation,
CureWM and the baseline receive 3,600 further
training steps.
CureWM uses all 145 training trajectories;
the baseline uses the 31 unperturbed trajectories.
Video losses have equal weight across training records.

The 71-failure diagnostic evaluates the adapted
model before repair.
Most examples subsequently enter repair training,
so the diagnostic is distinct from the held-out
repair evaluations.

\subsection{Prediction Scores and Scene Adaptation}
\label{app:hardware}

\paragraph{Camera inputs and score definitions.}
The hardware has external and wrist cameras,
but the model receives three inputs:
external, external, and wrist.
Distance calculations average over all three,
giving the external view twice the weight
of the wrist view.

Both action scores use
Equation~\ref{eq:video-margin}.
Visual optimism counts $s(a^-)>0$;
pair-ranking accuracy counts $s(a^+)>s(a^-)$;
and $\Delta_{\mathrm{SF}}$ averages their difference.
In the recorded outputs,
\texttt{fidelity\_plus} is distance to the successful
reference, not the signed score:
\[
s(a^+)
=
\texttt{fidelity\_plus\_cross}
-
\texttt{fidelity\_plus}.
\]

The two-reference failure detector uses $-s(a)$.
The success-reference-only detector uses
$d(\hat z_T(a),\phi(y_T^+))$.
Both require recorded reference outcomes.
Across the two evaluations, baseline-to-CureWM
AUROC is 0.66 to 0.96 with both references,
and 0.67 to 0.90 with the successful reference only.
Table~\ref{tab:real} reports two-reference AUROC;
the main-text 0.67-to-0.90 result uses one reference.

\paragraph{Rendering-quality screening.}
For evaluation pairs $\mathcal P$, we compute
\begin{equation}
\mathrm{SNR}
=
\frac{
|\mathcal P|^{-1}
\sum_{i\in\mathcal P}
d(\phi(y_{T,i}^+),\phi(y_{T,i}^-))
}{
|\mathcal P|^{-1}
\sum_{i\in\mathcal P}
d(\hat z_{T,i}(a_i^+),\phi(y_{T,i}^+))
}.
\label{eq:hardware-snr}
\end{equation}
The heuristic requires $\mathrm{SNR}>1$.
It compares separation between recorded outcomes
with prediction error on successful actions
and is a ratio of mean distances.
Passing does not guarantee valid inference.

A re-check of an earlier sponge evaluation found
SNR values of 0.73 and 0.52 and led to withdrawal
of its transfer interpretation.
The rule was subsequently fixed before the cube-task
and second cup-task evaluations, not before the
original sponge experiment.

\paragraph{Second hardware evaluation.}
Twenty new demonstrations produce 40 replays,
with two perturbation types per demonstration
at severity 1.0.
Insufficient grip, premature release, and carry slip
each fail on 10/10 replays;
wrist tilt succeeds on 10/10.

Before additional adaptation, SNR is
0.85/0.83/1.02/0.97 for the adapted model,
baseline, and CureWM models 1 and 2.
Three models fail the screen.
Their optimism rates are 90/90/10/17\%;
these measurements are excluded from the combined
repair results.

Ten demonstrations are then used for additional
scene adaptation and ten for evaluation, balanced
over start locations and perturbation types.
All models receive 350 steps on the same successful
demonstrations with unchanged normalization statistics.
Evaluation demonstrations and replays remain excluded.
The expected repair advantage and requirement to
pass the screen are recorded before this training.

After adaptation, SNR is 1.11/1.15/1.24/1.14.
Results on the 15 held-out failures appear in
Table~\ref{tab:real}.
Combining this evaluation with the first gives
30 pairs from 20 demonstrations.
Each model uses the adaptation corresponding
to its evaluation.

\paragraph{Successful-replay comparison.}
The additional baseline combines 31 unperturbed
trajectories with the same 24 successful wrist-tilt
replays used by CureWM.
Initialization and 3,600 training steps are matched.
First-evaluation optimism remains 15/15,
while mean $s(a^-)$ decreases from 0.184 to 0.138.
This does not reproduce CureWM's repair.
However, 55 versus 145 distinct trajectories
remain unmatched, so data content, quantity,
and diversity are not fully separated.

\subsection{Human Assessment}
\label{app:human}

Predictions for both actions in the first evaluation's
15 held-out pairs are generated by the adapted
released model and CureWM, producing 60 clips.
Each rater sees all 30 failing-action clips and
ten successful-action clips, five per model.
Three non-author volunteers unfamiliar with the
project classify clips as successful, failed,
or unclear.
Raters were informed of the procedure and research
use of their ratings and voluntarily agreed
to participate.
Only generated videos are shown, without recorded
reference outcomes.
Clip identifiers are hashed, order is independently
shuffled for each rater, and the interface does
not load the answer key.
All raters complete 40 clips and none selects unclear.
The fine-tuning baseline is not evaluated. 
Table~\ref{tab:human} summarizes the majority judgments.

\begin{table}[htbp]
\centering
\small
\caption{Majority judgments from three raters.
The diagnostic classifies predictions as success-like
when $s(a)>0$.}
\label{tab:human}
\setlength{\tabcolsep}{5pt}
\begin{tabular}{@{}llrr@{}}
\toprule
Executed action & Model
& Human: successful & Diagnostic: success-like \\
\midrule
Failing & Adapted & 8/15 & 13/15 \\
Failing & CureWM & 0/15 & 6/15 \\
Successful & Adapted & 3/5 & --- \\
Successful & CureWM & 5/5 & --- \\
\bottomrule
\end{tabular}
\end{table}

Individual raters classify 8, 7, and 8 adapted-model
failing-action predictions as successful, versus
zero for CureWM.
Pairwise Cohen's $\kappa$ values are 0.95, 0.90,
and 0.84.
Human judgments agree on the direction of improvement,
but the distance diagnostic more frequently
classifies predictions as successful.
This small evaluation does not calibrate the separate
71-failure diagnostic or establish preservation
of successful predictions.

\subsection{Cube Picking and Cross-Task Transfer}
\label{app:cube}

The robot picks a colored cube among distractors.
Collection contains 112 episodes across 29 scenes,
including 83 verified failures.
An operator places objects between scenes,
and scripts execute the trajectories.
Gripper telemetry distinguishes a held 2\,cm cube
(approximately 0.70) from an empty close
(approximately 1.00), using threshold 0.88.
For \texttt{wrong\_target}, target identity additionally
uses known placement geometry and the commanded
distractor target, with operator review of anomalies.
Ten scenes containing 29 counterfactual pairs
are reserved for evaluation and excluded from
cube-scene adaptation and repair.
Split membership is derived from the training
split file.
The baseline and task-specific CureWM start from
the same cube-adapted checkpoint.
An initial evaluation at SNR 0.84 is excluded;
additional scene adaptation raises SNR to 1.17.
We evaluate both models at 300, 600, 1,200, 1,800,
2,400, 3,000, and 3,600 training steps, giving
fourteen model--checkpoint evaluations.
The primary comparison uses 3,600 steps.
All fourteen model--checkpoint evaluations pass
the screen, with SNR 1.06--1.18. Table~\ref{tab:cube-results} compares task-specific
repair with transfer from cup-task repair.

\begin{table}[htbp]
\centering
\small
\caption{Cube picking on 29 held-out pairs from ten scenes.
Differences are baseline minus repaired-model mean
$s(a^-)$; positive values indicate lower failure-action
scores after repair.}
\label{tab:cube-results}
\setlength{\tabcolsep}{8pt}
\begin{tabular}{@{}lr@{}}
\toprule
Repair setting & Failure-score difference \\
\midrule
Cube-task repair & $+0.025$ \\
Cup-task repair before cube adaptation & $-0.007$ \\
\bottomrule
\end{tabular}
\end{table}

Across checkpoints from 600 steps onward,
detection AUROC ranges from 0.927 to 0.976
for CureWM and from 0.748 to 0.924 for the baseline.
These are changes along training trajectories,
not variation across independently trained models.
The cup-to-cube comparison does not establish
beneficial transfer; the positive result uses
cube-specific counterfactual training data.

\end{document}